# The chemical space of terpenes: insights from data science and AI

Morteza Hosseini, David M. Pereira*

REQUIMTE/LAQV, Laboratório de Farmacognosia, Departamento de Química, Faculdade de Farmácia, Universidade do Porto, Rua de Jorge Viterbo Ferreira, Nº 228, 4050-313 Porto, Portugal

[*]Corresponding author

*Tel.: +351 220 428 655;*

*E-mail addresses:* dpereira@ff.up.pt (D.M.P.)

# Abstract

Terpenes are a widespread class of natural products with significant chemical and biological diversity and many of these molecules have already made their way into medicines. Given the thousands of molecules already described, the full characterization of this chemical space can be a challenging task when relying in classical approaches.

In this work we employ a data science-based approach to identify, compile and characterize the diversity of terpenes currently known in a systematic way. We worked with a natural product database, COCONUT, from which we extracted information for nearly 60000 terpenes. For these molecules, we conducted a subclass-by-subclass analysis in which we highlight several chemical and physical properties relevant to several fields, such as natural products chemistry, medicinal chemistry and drug discovery, among others.

We were also interested in assessing the potential of this data for clustering and classification tasks. For clustering, we have applied and compared $k$-means with agglomerative clustering, both to the original data and following a step of dimensionality reduction. To this end, PCA, FastICA, Kernel PCA, t-SNE and UMAP were used and benchmarked.

We also employed a number of methods for the purpose of classifying terpene subclasses using their physico-chemical descriptors. Light gradient boosting machine, $k$-nearest neighbors, random forests, Gaussian naïve Bayes and Multilayer perceptron, with the best-performing algorithms yielding accuracy, F1 score, precision and other metrics all over 0.9, thus showing the capabilities of these approaches for the classification of terpene subclasses.

# Introduction

The natural products chemical space is a fascinating set of hundreds of thousands of molecules that are not only remarkably interesting from a strictly chemical point of view, but also owing to the diverse and impressive set of biological properties that many of these molecules possess. The importance of this chemical space is further evidenced by the significant number of such molecules are currently used as medicines in human and veterinary medicine.

From a phytochemical and pharmacognostic point of view, natural molecules are grouped into different families on the grounds of their biosynthetic origin or sometimes due to their shared chemical traits.

Many families of natural molecules are of interest to human health, not only given their role in medicine but also given their application in nutrition or cosmetics. Among such families, terpenes are a diverse set of compounds that have paved their way not only as medicines (artemisinin, taxol, among others) but also in industries as diversified as foodstuffs (carotenoids as coloring agents), flavors (menthol, limonene, pinene) and preservatives (eugenol).

Given the tens of thousands of terpenes known, they are frequently grouped on the grounds of their number of carbons, which in turn reflect their biosynthetic approach. Briefly, depending on the number of $C_5$ isoprene units, terpenes can be monoterpenes ($C_{10}$), sesquiterpenes ($C_{15}$), diterpenes ($C_{20}$) triterpenes ($C_{30}$) and so on. Some of these molecules can also be lactones or bear sugars, thus being routinely classified as terpene lactones or terpene glycosides, respectively.

Like other classes of natural products, the fast pace at which new molecules are described makes it increasingly difficult to continue to manually curate and study the chemical descriptors of each molecule individually. To this end, data science-based approaches are needed, as they allow to organize, interpret and filter huge amounts of data and, sometimes, highlight relationships that would otherwise eclipse human attention.

As frequently postulated, the quality of data in data science-centered analysis is of paramount importance. Fortunately, the digitalization of information has enabled data science-based frameworks to study the natural products chemical space.

In this work, we employ a data science and artificial intelligence approach to the largest and most complete database of natural products to date, the COCONUT database. The latest reference we could find stated that over 55000 terpenes are known (Guimarães et al., 2014), and now we extracted 59000 terpenes from this resource and employed a data science approach to extract, describe and interpolate data from this data source. The result is a compendium of information on terpenes, on a subclass-by-subclass basis, that can be a valuable resource for researchers, teachers and students of this exciting branch of natural products chemistry and pharmacognosy.

## Methods

### Data collection

We started with the COCONUT database available at https://coconut.naturalproducts.net. The dataset had the original shape of 401624 entries in 1554 columns. Initial data assessment showed that in most of the columns, more than 70% of entries were NULL. We dropped those columns. Also, there were a few unique identifiers, such as "_id" and "coconut_id", which had no predictive or informative value. Such variables were dropped, thus originating the dataset used, with 401624 entries and 45 variables, which are described in Supplementary Table 1. We then selected only the entries that belonged to the SuperClass "Lipids and lipid-like molecules" (Supplementary Table 2) and further filtered only the molecules that belonged to one of the following SubClasses: "Diterpenoids", "Sesquiterpenoids", "Monoterpenoids", "Polyterpenoids", "Sesquaterpenoids", "Sesterterpenoids", "Terpene glycosides", "Terpene lactones" and "Triterpenoids". After this selection, the resulting dataset included 59833 molecules.

## Data cleansing

To preprocess the data, multiple steps were taken. First, we handled categorical features, including "textTaxa", "bcutDescriptor", "chemicalClass", "chemicalSubClass", "chemicalSuperClass" and "directParentClassification". To encode the "textTaxa" feature, we created four new columns for "plants", "marine", "bacteria" and "fungi" taxonomy in a way that for each molecule having either of those values, a 1 was inserted to the corresponding column, and 0 otherwise. The "bcutDescriptor" feature included arrays of six float numbers; we split and expanded those into six separate columns. Since all terpenes share the same "chemicalClass" and "chemicalSuperClass", we did not consider them for further processes. "chemicalSubClass" was the target, so we did not encode it. The "directParentClassification" feature, containing 111 values, were simply encoded by the integers 0 to 110.

For further processing, we split the dataset into 75% training and 25% testing data. Next, we imputed the data using median strategy, i.e., replaced missing values with the median of not-null training values in each column. Then, in order to scale the data, we standardized it by centering the data around the mean (removing the mean) with a unit variance. Note that the features with binary values, such as "contains_sugar", do not need to be scaled.

## Clustering

For clustering, we applied *k*-means and agglomerative clustering methods on the dataset, as a whole, and also on the dimensionality reduced form of it. *k*-means is a widely used method that partitions the data into *k* clusters in a way that each datapoint belongs to the cluster with the closest centroid (cluster center). Agglomerative clustering, another popular method, is a "bottom-up" approach to perform hierarchical clustering, in which starting from each datapoint, that forms one cluster, pairs of clusters are recursively merged based on linkage distance. Clustering has been widely used in the field of medicine and biomedical sciences, being used for applications such as selecting new candidate drugs for lung cancer (Lu et al., 2016), molecular descriptor analysis (Madugula et al., 2021) or clustering the gene profiles of distinct types of cancer (Pawar et al., 2020).

Several dimensionality reduction algorithms were employed for our purpose, namely PCA, FastICA (Hyvärinen and Oja, 2000), Kernel PCA (Schölkopf et al., 1998), t-SNE (Van der Maaten and Hinton, 2008) and UMAP (McInnes et al., 2018). PCA (principal component analysis) changes the basis of data by computing principal components in order to project the high-dimensional data into a lower dimensional space. FastICA (fast independent component analysis) is an efficient algorithm that finds an orthogonal rotation of prewhitened data, with the statistical independence assumption. Kernel PCA (kernel principal component analysis) is as an extension of PCA that employs kernels, such as RBF (radial basis function), to reduce the dimensionality of data, non-linearly. t-SNE (t-distributed stochastic neighbor embedding) is a non-linear technique that employs stochastic neighbor embedding to project each datapoint in a high-dimensional space into a location in a 2- or 3-dimensional space. UMAP (uniform manifold approximation and projection) is a manifold learning method that is based on Riemannian geometry and algebraic topology; its output is visually similar to t-SNE, but it does not have computational limitations on embedding dimension. In the field of biomedical sciences PCA is the most widely used method (Ringnér, 2008), however UMAP has been receiving significant attention since its introduction (Becht et al., 2019).

To carry out the above-mentioned methods, we used "scikit-learn" and "umap" Python libraries, that can be accessed at https://github.com/scikit-learn/scikit-learn and https://github.com/lmcinnes/umap, respectively. Also, the parameters used are listed in Supplementary Table 3. Note that we performed clustering on the data belonging to "Triterpenoids", "Diterpenoids" and "Monoterpenoids" subclasses.

## Classification

We carried out multiple methods, namely Light gradient boosting machine (LightGBM) (Ke et al., 2017), $k$-nearest neighbors ($k$NN) (Fix and Hodges, 1989), random forests (Ho, 1995), Gaussian naïve Bayes (Breiman, 1996) and Multilayer perceptron (MLP), to classify six subclasses of "Monoterpenes". "Triterpenes", "Diterpenes", "Sesquiterpenes", "Terpene lactones" and "Terpene glycosides". LightGBM is a distributed gradient boosting framework that works based on decision tree algorithms. In the $k$NN method, the class membership of an object is determined by the most common class between its *k* nearest neighbors. Random forests are an ensemble learning method in which different decision tree classifiers are fit on various sub-samples of the data. Gaussian naïve Bayes is based on the Bayes theorem and follows the Gaussian normal distribution. Multilayer perceptron is a feedforward neural network with at least three layers of input, hidden and output, that uses backpropagation for training and uses a non-linear activation function. Note that for the implementation, we used "scikit-learn" and "lightgbm" ([https://github.com/microsoft/LightGBM](https://github.com/microsoft/LightGBM)) Python libraries.

## Results

Our primary source of data was the COCONUT database, created and curated by the Institute for Inorganic and Analytical Chemistry, Friedrich-Schiller University, Germany (Sorokina et al., 2021). This resource, first made available in 2021, is an online repository of nearly half a million natural products annotated with several physico-chemical and topographical information and compiles the information of over 50 databases of natural molecules.

Starting from the COCONUT database (400k+ molecules), we initially created a subset comprising solely the chemical SuperClass "Lipids and lipid-like molecules", which comprises all molecules of lipidic nature, in which terpenes are included. Among this subset (99696 molecules, distributed according to Supplementary Table 1), we further selected the classes "Monoterpenes, "Sesquiterpenes", "Diterpenes", "Sesterterpenes", "Triterpenes", "Polyterpenes", "Sesquaterpenes", "Terpene glycosides" and "Terpene lactones", in order to include all terpene classes. This yielded the final dataset, which included 59833 molecules. We acknowledge that many molecules in the "Terpene glycosides" and "Terpene lactones" subclasses can be, from a phytochemical point of view, part of any of the remaining subclasses. However, for the sake of clarity and access to as many molecules as possible, we followed the hierarchical criteria enforced by COCONUT.

**Figure 1** depicts the distribution of the molecules in the dataset across different terpene classes. As shown, the major classes were triterpenes (22.1%, 13245 molecules), followed by diterpenes (19.7%, 11814 molecules), and the least expressive class was polyterpenes (0.1%, 48 molecules).

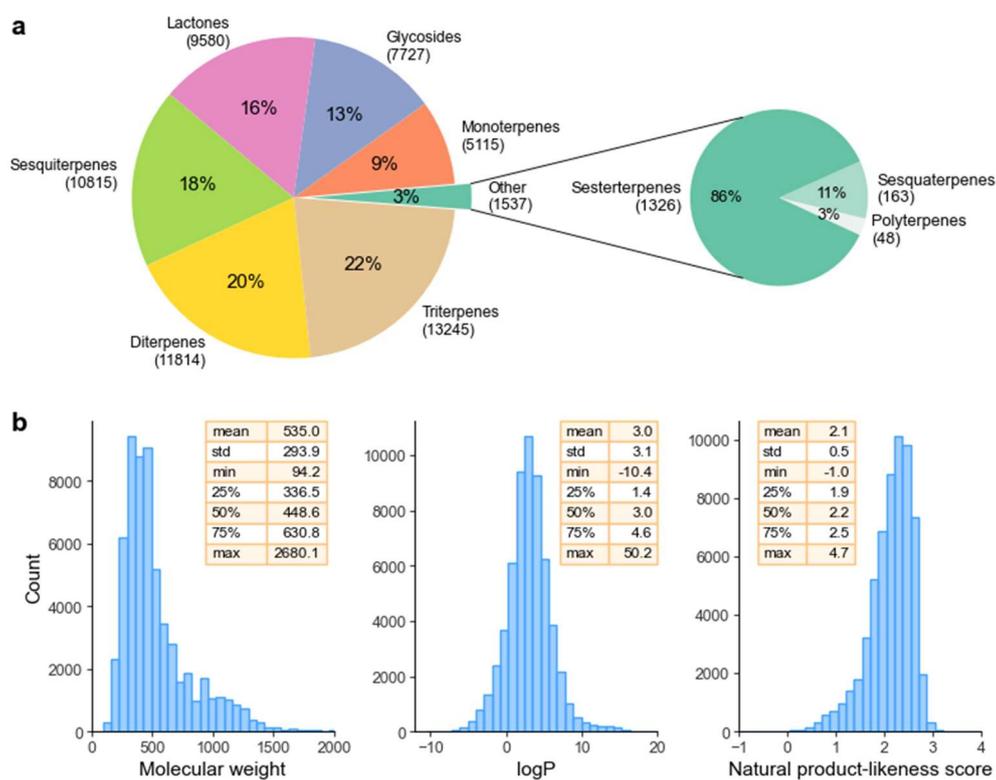

**Figure 1.** (a) Overview of absolute and relative counts of terpene subclasses. 1: Sesterpenes; 2: Other classes [Sesquaterpenes, 163, 0.2%; Polyterpenes, 48, 0.1%). (b) Distribution of molecular weight, natural product-likeness score and logP for all terpenes in the dataset. For the sake of clarity, we have chosen only molecular weights up to 2000 and -12 < logP < 20, with only 0.2% and 0.1% of the molecules being suppressed, respectively.

Before advancing, it is important to briefly review the organization of COCONUT in terms of chemical ontology. After a registry is created, several parameters are automatically calculated based on well-established algorithms. For example, in the case of the chemical classification, Classyfire is used (Djoumbou Feunang et al., 2016). This is a remarkable resource, as this is a purely structure-based chemical taxonomy software that uses chemical structures and structural features as inputs to assign a given molecule to a hierarchical-based taxonomy. This new chemical taxonomy consists of up to 11 different levels (including Kingdom, SuperClass, Class, SubClass, etc.) with each of the categories defined by unambiguous and computable structural rules.

This approach is both powerful and dangerous in the case of the natural products chemical space. On the one hand, we are able to easily classify hundreds of thousands of molecules based on clearly established rules, instead of having to rely in biosynthetic studies. As an example, when we run taxol into Classifyre, we get the classification: Kingdom: Organic compounds > Superclass: Lipids and lipid-like > Class: Prenol Lipids > Subclass: Diterpene > Classification: Taxanes and derivatives (**Figure2**). From a phytochemical point of view, one would not usually include terpenes in the class of lipids (older textbooks would, however) but it is true they are of lipidic nature, so there is no surprise here. The most important levels of taxonomical classification for a natural products chemist are Subclass: Diterpenes and Parent: Taxanes. This works fine as we can get both the natural product family and its nucleus/immediate family. To be able to have this degree of information even in cases where biosynthetic studies are not yet available is remarkable.

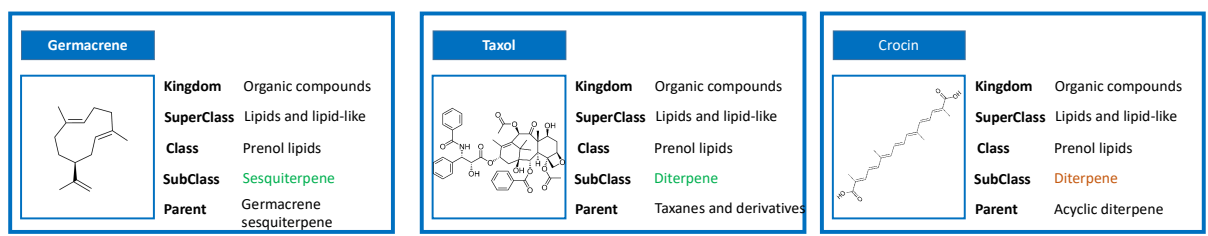

**Figure 2.** Chemical classification of germacrene, taxol and crocin according to ClassyFire.

On the other hand, the disadvantage of this rule-based classification is that natural products are a wonderful source of rule breakers and deviants. Take for instance crocin, from *Crocus sativus*. From a phytochemical point of view crocin is a (di)apocarotenoid, as its loss of isoprene units from both ends renders it a chemical entity with $C_{20}$. Naturally, a rule-based algorithm has no way to know that from a biosynthetically point of view this molecule is a carotenoid (typically $C_{40}$) that has lost carbons. As so, it is classified as "acyclic diterpenoid" (compounds comprising four consecutive isoprene units that do not contain a cycle, **Figure 2**).

Nevertheless, despite these limitations, it is helpful to have a tool to quickly classify millions of molecules, with the advances in technology allowing to fine-tune the rules towards increasingly accurate performances. However, some caution must be taken to structures that may be incorrectly classified. In **Figure 3** we present the different levels of chemical taxonomy for the dataset used in this work.

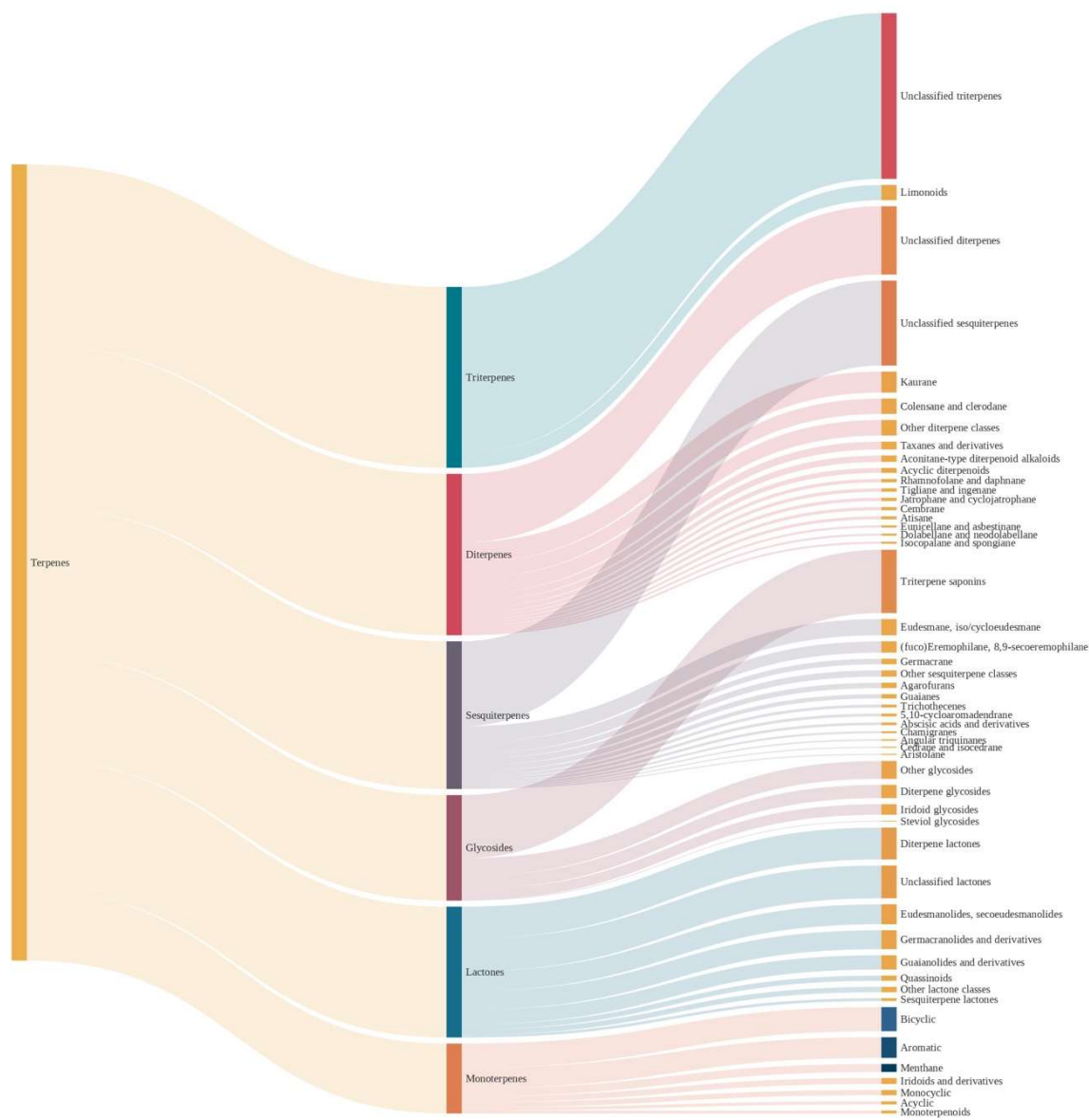

**Figure 3.** Classification of terpenes in the COCONUT database. The second level of classification corresponds to classical phytochemical subclasses (monoterpenes, diterpenes, sesquiterpenes, among others), while the last level corresponds to characteristic nuclei or subset.

Considering all terpenes studied here, the average molecular weight was 534.9 Da (min: 94.2, nortricyclene and max 2680.1, palytoxin, **Figure 4**). Nortricyclene does not comply with the standard monoterpene definition given its odd number of carbons, 7, however this classification arises from its structural relation to tricyclene ($C_{10}H_{16}$, **Figure 4**). Palytoxin, from *Palythoa* corals, is one of the largest non-polymeric natural products described and one of the most poisonous non-protein molecules known, second only to maitotoxin in terms of mice toxicity (Sud et al., 2013). It is a terpene glycoside and displays 8 double bonds, 40 hydroxyl groups and 64 chiral centers, which gives rise to over $10^{21}$ possible stereoisomers.

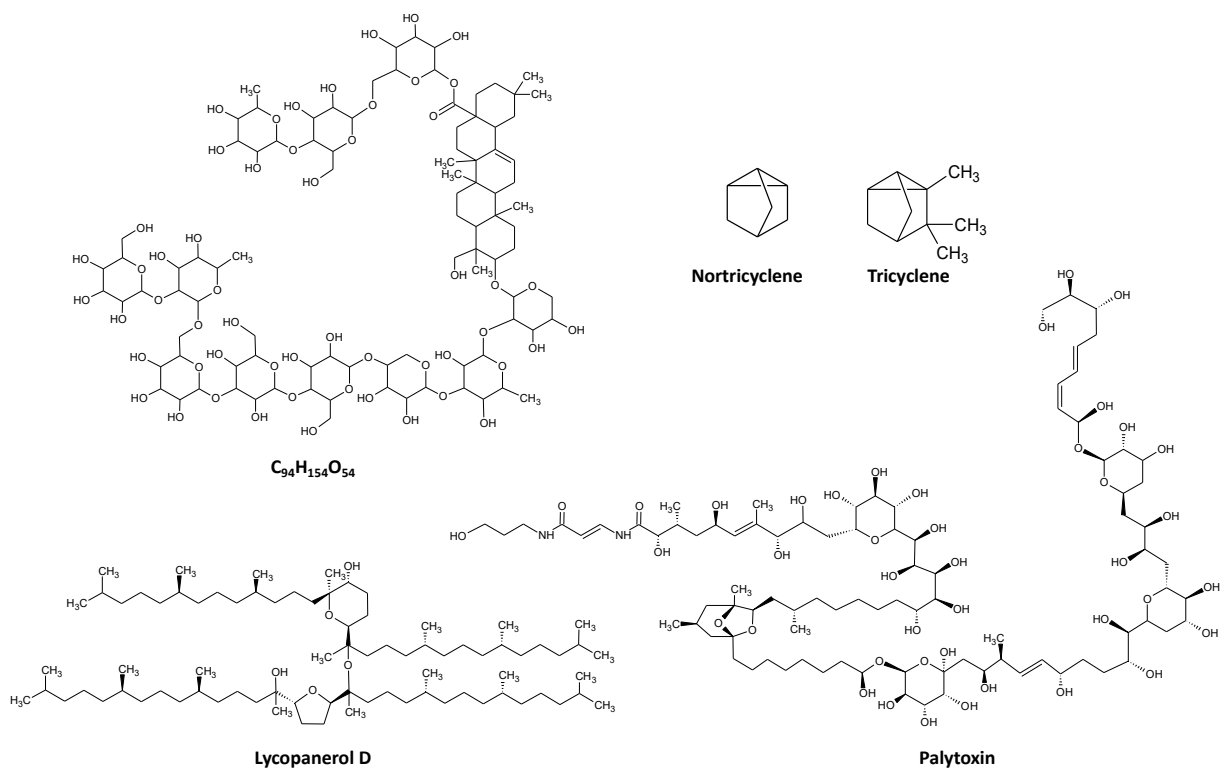

**Figure 4.** Structure of the terpenes that occupy the lower and upper range of the MW and logP in the terpene family. Nortricyclene is the smallest molecule in the database (94.15 Da), while palytoxin is the one with higher MW (2680.14 Da). The unnamed molecule (no common name) is the entry with the lowest logP (-10.4) while lycopanerol D presents the highest calculated logP (50.2).

The distribution of MW across all terpenes is presented in **Figure 1**. From a phytochemical point of view, the different terpene subclasses are grouped based on their biosynthesis and, consequently, the number of isoprene units in their structure. For this reason, it is obvious that marked differences in molecular weight can be found across distinct subclasses. For the sake of compiling the molecular weight intervals in each terpene class, **Figure 5**a shows molecular weight distribution as a function of subclass. The average molecular weight increases with the increasing number of carbons in the monoterpene < sesquiterpene < diterpene < triterpene order. Terpene glycosides present the widest range of molecular weight, an expected consequence of the occurrence of different number of sugars and their diverse identity.

The partition coefficient (P), which describes the suitability of a neutral molecule to dissolve in immiscible biphasic systems comprised of lipids and water, is a pivotal characteristic that has a marked impact in the "drugability" of a compound, as extreme values, either too negative or too positive can hinder its adequate pharmacokinetics in the human body. For example, a study comprising a dataset of four leading pharmaceutical companies showed that among 812 molecules studied as drug candidates, the average logP was 3.2. As shown in **Figure 1**, the average logP of terpenes is approximately 3, with a molecule with formula $C_{94}H_{154}O_{54}$ (no common name) and the tetraterpenoid ether lycopanerol H being the molecules in the lower and upper limit (-10.3 and 50.2, respectively, **Figure 4**). As shown in the class-by-class analysis (**Figure 5**b), the average value increased in the monoterpene < sesquiterpene < diterpene < triterpene order, the latter being the subclass of terpenes with the highest average logP, 4.6. Terpene glycosides was the only class presenting a negative average logP, as expected given their heterosidic nature that greatly increases their polarity.

The natural product-likeness score is a descriptor that takes into account several structural and topographical features and tries to quantify the "degree of naturalness" of a given molecule (Sorokina and Steinbeck, 2019). Naturally occurring molecules, owing to their stereochemical complexity and diversified ring systems, usually afford higher values whereas compounds arising from synthetic chemistry display lower values. As shown in **Figure 5**c, the average value for terpenes was 2.1. Monoterpenes, the simplest class from a structural point of view, had the lowest average natural product-likeness score, 1.5, owing to their simpler chemical structures. This value increases with the chemical complexity of the subclasses in the monoterpene > sesquiterpene > diterpene > triterpene order.

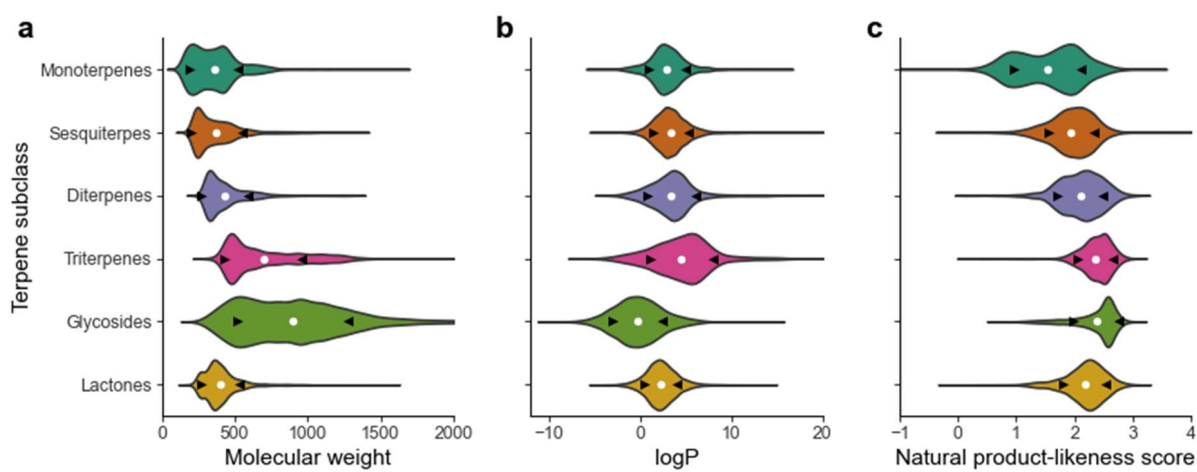

**Figure 5.** Distribution of (a) molecular weight, (b) logP and (c) natural product-likeness score across different terpenes subclasses. White mark (●) represents the mean. Left black arrow (▶) represents the mean - std and the right arrow (◀) the mean + std.

Several terpenes are currently being used in the drug arsenal against a number of diseases. This includes, for example, paclitaxel for cancer, artemisinin for malaria or manoalide for inflammatory processes. In the process of pharmaceutical development, it is important to take into account the "drug-likeness" of a given molecule in order to identify unlikely candidates as upstream in the development process as possible. One of the most widely spread set of "rules of thumb" are the Lipinski's rule of five (Benet et al., 2016), which postulate that ideal drugs would have number of hydrogen bond donors $\leq 5$, number of hydrogen bond acceptors $\leq 10$, molecular mass $\leq 500$ Da and logP $\leq 5$. Although this set of anecdotal rules have been postulated a long time ago, most of its basis is still widely used today: in a study with over 800 drug candidates, only 24% had a molecular weight > 500 Da and only 15% of the candidates had a logP > 5. Notably, only 7.6% of all the molecules violated both rules simultaneously (Waring et al., 2015).

We also investigated the rate of violation of the terpenes chemical space in terms of number of hydrogen bond donors and acceptors. As shown in **Figure 6**, 78.7% of all terpenes had ten hydrogen bond acceptors or less, while 81% had five hydrogen bond donors or less, thus complying with Lipinski's rule of five. When studying these parameters on a subclass level, we see that in a general way, monoterpenes, sesquiterpenes, diterpenes and terpene lactones all present over 85% of molecules with ten or less hydrogen bond acceptors and 70% of molecules with five or less hydrogen bond donors. Triterpenes are slightly different, as they exhibit 70% and 59% for the same parameters, respectively. The most distinct set of molecules are terpene glycosides, in which only 18.7% of molecules observe these rules.

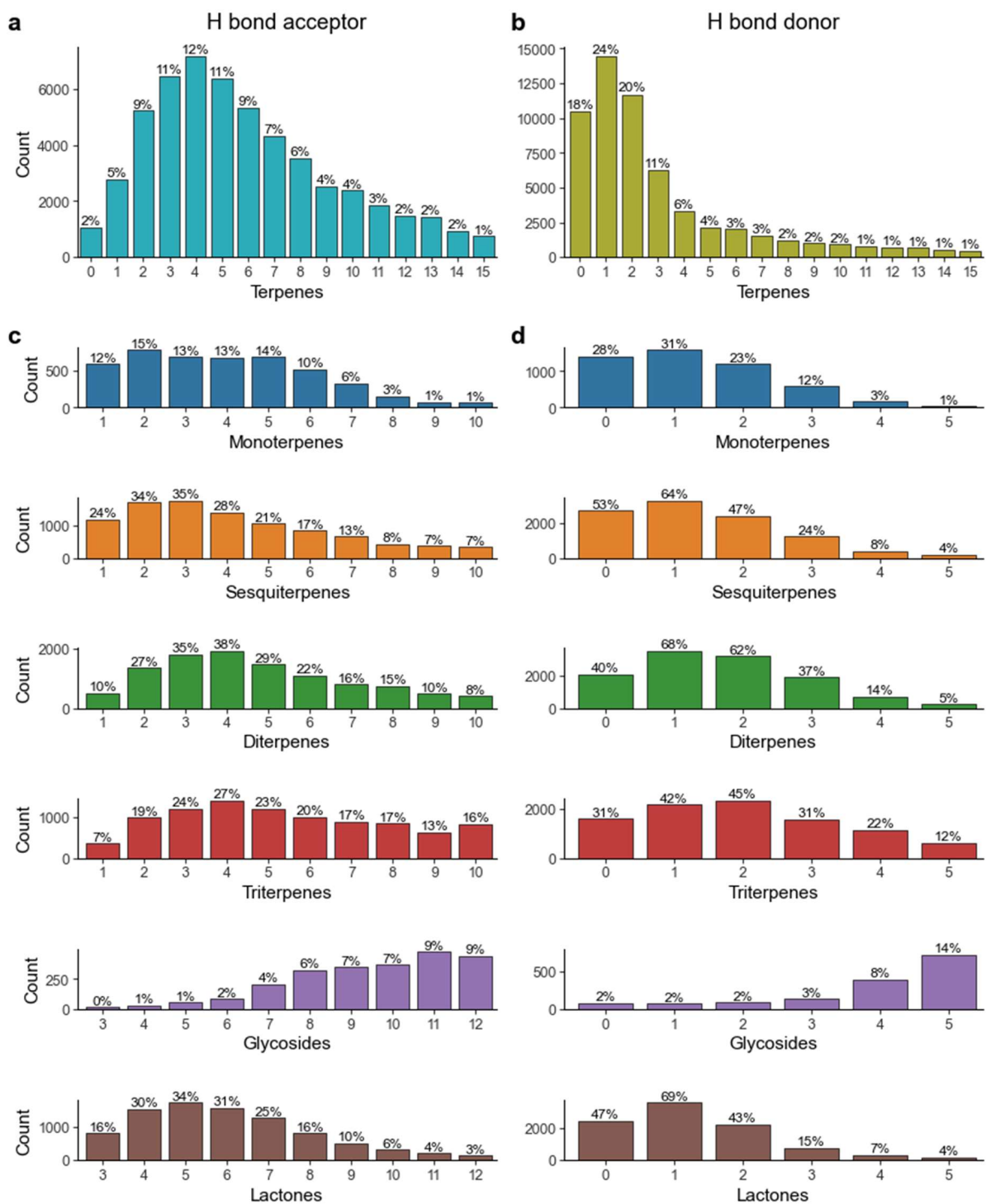

**Figure 6.** Number of hydrogen bond (a) acceptors and (b) donors in the terpene chemical space and on a subclass basis for (c) acceptors and (d) donors.

After this, we were interested in quantifying how many molecules violated any of the postulated rules. As shown in **Figure 7**, 41.6% of terpenes do not present any violation. While the most common deviation is 1 violation (22.8%), it is of note that there is an almost equivalent count of molecules with 2, 3 and 4 violations, which cumulatively account for 35% of all molecules. We further investigated the number of violations across different terpenes subclasses. As seen in **Figure 7**, there are remarkable differences. Monoterpenes and sesquiterpenes present a similar and nearly superimposable profile and distribution of number of violations. Triterpenes are unique in the way they present the lowest percentage of compounds with 0 violations, only 5.2%, while nearly 60% present 1 or 2 violations. Equally unique are terpene glycosides, which include over 50% of molecules with 4 violations, a unique trait, considering that the second class in this category is sesquiterpenes, with only 1.9%. Likewise, terpene lactones are a rather particular class, as nearly 80% of the molecules in this class present no violations to Lipinski's rule of five. The results presented so far provide a valuable source of information that can be used to select the terpene subclasses that present a chemical profile more compatible to future drug development frameworks.

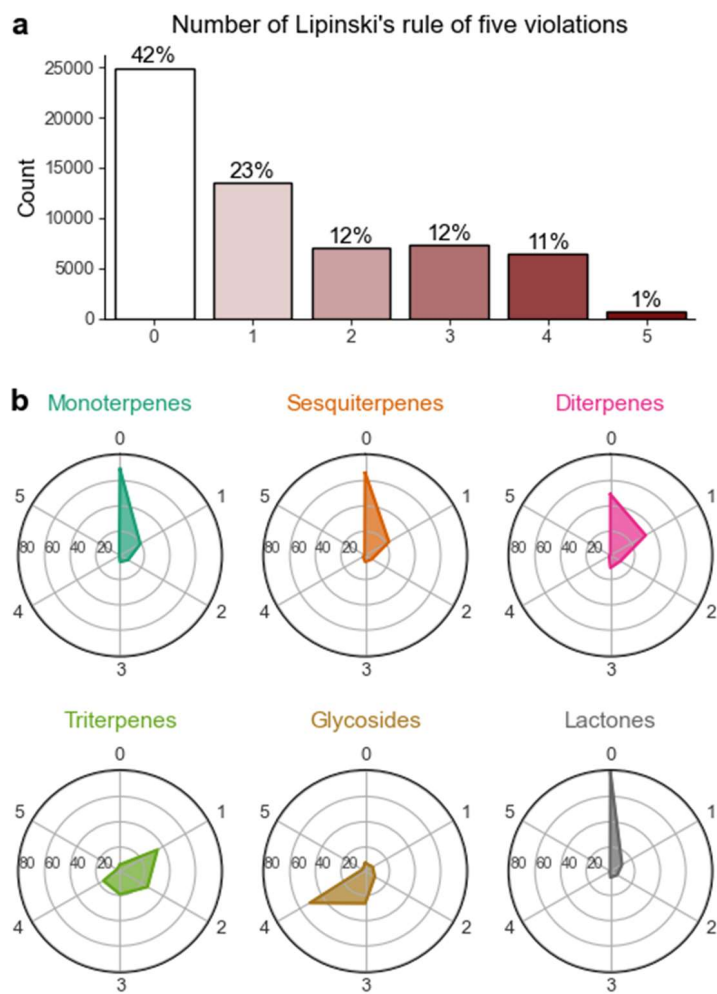

**Figure 7.** (a) Number of violations to the Lipinski's rule of five in the terpene chemical space, for all terpenes, and (b) across distinct subclasses; outer values correspond to the number of violations and inner values the percentage of a given number of violations.

## Clustering the terpene chemical space

In high dimensional data, clustering can be an effective method to make sense of heterogeneous data. In the case of chemical information, it can be used to identify the most important physico-chemical parameters that allows to group the different families of molecules. In the specific case of terpenes, this can be useful where a new chemical entity is found and its sorting in a given class is challenging owing to inexistent biosynthetical data, mostly in the case of molecules with an odd number of carbons.

In the case of terpenes, we were interested in studying if their physico-chemical parameters could per se help position a given molecule in a particular subclass. To this end, and given the high dimensionality of data, we employed several dimensionality reduction methods, namely PCA, UMAP, t-SNE, FastICA and Kernel PCA. After this, we tried to cluster the data, both the original data along with the reduced form, by *k*-means and hierarchical clustering and benchmarked the results.

To evaluate the quality of clustering, we considered the following metrics. Homogeneity score (shorthand: Homo) measures if each cluster includes only the datapoints that are members of a single class. Completeness score (shorthand: Compl) shows if all datapoints of a given class are members of the same cluster. The harmonic mean between the homogeneity and completeness scores is computed by v-measure (shorthand: V-meas) (Rosenberg and Hirschberg, 2007). The Rand index (Rand, 1971) shows the similarity between two label assignments and the adjusted Rand index (shorthand: ARI) is a version of the Rand index that corrects for chance. The agreement between two label assignments, ignoring permutations, is computed by mutual information. Here, we used adjusted mutual information (shorthand: AMI), which is normalized against chance version of it. The above-mentioned measures fall within the range 0 to 1, with score 1 meaning perfectly labelling. The silhouette coefficient (Rousseeuw, 1987), however, ranges from $-1$ to 1 and measures how similar a datapoint is to its own cluster rather than other clusters. Higher silhouette values show that a datapoint is highly similar to its own cluster and poorly similar to neighboring clusters. In addition to mentioned metrics, we measured the execution time of different methods running on Apple M1 CPU.

Supplementary Tables 4 and 5 demonstrate, in detail, the results of carrying out *k*-means and agglomerative clustering on imbalanced and balanced datasets, with and without applying dimensionality reduction methods with the parameters shown in Supplementary Table 3. Of those results, the best ones are shown in Table 1. In the case of *k*-means on imbalanced data, PCA provides metric values ≤ 0.33 with 11 principal components, which are low-performing values, as they should be as close to 1 as possible. When compared to PCA, UMAP (n_neighbors=45, min_dist=0.1, Supplementary Table 3) did not provide any substantial increase in performance, except for silhouette which yielded an increase of 0.21. t-SNE did not provide any improvement over UMAP, despite the multiple settings tried; the same was true for FastICA and Kernal PCA, except for the silhouette metric. Kernal PCA afforded the best silhouette, 0.50, which is still low and hence not suitable for our clustering purposes. Considering these results, we wondered if the reason for this poor performance could be related to some degree of imbalance in the data. To account for this, we applied random oversampling (ROS) for data balancing, after which all algorithms were run once more. Note that since it was important for us not to generate synthetic data, we did not employ synthetic sampling approaches. As shown in the table, data balancing did not improve the results; the three subclasses we fed to the clustering algorithm, i.e., "Triterpenoids", "Diterpenoids" and "Monoterpenoids", had the occurrences of 13245, 11814 and 5115, respectively.

The same procedure as above was repeated by hierarchical (agglomerative) clustering. Table 1 shows that applying UMAP (UMAP7), before clustering, provides the best results when we cluster on imbalanced data. These results are almost the same as the ones obtained by UMAP7 plus *k*-means, with the difference that the latter is 70% faster. Similar to before, data balancing could not help improving the results. Therefore, the best results are obtained by first applying UMAP7 to reduce dimensionality of the data (the original imbalanced one) and then, *k*-means to cluster it. Note that even these methods could not provide a satisfactory result in clustering the three target subclasses.

**Table 1.** Clustering the original terpenes data along with dimensionality reduced form of it, in both imbalanced and balanced forms. Parameters are described in Supplementary Table 3.

| Dim reduce | Time (s) | Homo | Compl | V-meas | ARI | AMI | Silhouette |
|---|---|---|---|---|---|---|---|
| ***k*-means on imbalanced data** | | | | | | | |
| original | 0* | 0.30 | 0.32 | 0.31 | 0.23 | 0.31 | 0.19 |
| PCA0 | 0* | 0.32 | 0.33 | 0.33 | 0.25 | 0.33 | 0.21 |
| UMAP7 | 19 | 0.39 | 0.40 | 0.40 | 0.32 | 0.40 | 0.42 |
| TSNE6 | 45 | 0.21 | 0.20 | 0.21 | 0.19 | 0.21 | 0.39 |
| FastICA | 0* | 0.23 | 0.25 | 0.24 | 0.16 | 0.24 | 0.46 |
| Kernel PCA | 72 | 0.24 | 0.23 | 0.23 | 0.20 | 0.23 | 0.50 |
| ***k*-means on balanced data** | | | | | | | |
| original | 0* | 0.27 | 0.31 | 0.29 | 0.25 | 0.29 | 0.21 |
| PCA1 | 0* | 0.27 | 0.30 | 0.29 | 0.24 | 0.29 | 0.40 |
| UMAP7 | 26 | 0.11 | 0.11 | 0.11 | 0.12 | 0.11 | 0.36 |
| TSNE6 | 65 | 0.24 | 0.24 | 0.24 | 0.24 | 0.24 | 0.38 |
| FastICA | 0* | 0.19 | 0.22 | 0.20 | 0.13 | 0.20 | 0.47 |
| Kernel PCA | 1353 | 0.25 | 0.25 | 0.25 | 0.24 | 0.25 | 0.45 |
| **Agglomerative clustering on imbalanced data** | | | | | | | |
| original | 23 | 0.32 | 0.32 | 0.32 | 0.27 | 0.32 | 0.13 |
| PCA0 | 16 | 0.32 | 0.32 | 0.32 | 0.28 | 0.32 | 0.16 |
| UMAP7 | 32 | 0.40 | 0.40 | 0.40 | 0.34 | 0.40 | 0.41 |
| TSNE8 | 127 | 0.21 | 0.20 | 0.21 | 0.24 | 0.21 | 0.32 |
| FastICA | 11 | 0.21 | 0.20 | 0.20 | 0.18 | 0.20 | 0.34 |
| Kernel PCA | 83 | 0.17 | 0.17 | 0.17 | 0.13 | 0.17 | 0.40 |
| **Agglomerative clustering on balanced data** | | | | | | | |
| original | 56 | 0.15 | 0.17 | 0.16 | 0.11 | 0.16 | 0.21 |
| PCA0 | 37 | 0.27 | 0.32 | 0.29 | 0.25 | 0.29 | 0.22 |
| UMAP7 | 51 | 0.10 | 0.11 | 0.10 | 0.10 | 0.10 | 0.34 |
| TSNE6 | 93 | 0.30 | 0.33 | 0.32 | 0.26 | 0.32 | 0.32 |
| FastICA | | 26 | 0.17 | 0.20 | 0.18 | 0.12 | 0.18 | 0.41 |
| Kernel PCA | | 1381 | 0.28 | 0.30 | 0.29 | 0.28 | 0.29 | 0.40 |

*Less than 1 second.

## Classification

We were also interested in assessing the suitability of this data for classification purposes. This is an important task, as it may help classify novel molecules to a given terpene subclass in cases where biosynthetic studies are not yet available. We fitted several classification methods, namely LightGBM, $k$NN, random forests, Gaussian naïve Bayes and MLP, with their default parameters in scikit-learn on the terpenes training data, that was obtained after data cleansing described in the "Methods" section; then, validated them by 5-fold cross-validation technique.

To evaluate the classification methods, we measured the following metrics which range from 0 to 1, with the values closer to 1 showing better performance. Considering the fact that the target feature, i.e., "chemicalSubClass", is imbalanced, we used the weighted version of the metrics. Precision shows if the classifier is able to not label negative samples as positive. Recall shows if the classifier is able to find all the positive samples. F1 score is calculated by taking the harmonic mean of the precision and recall. Balanced accuracy is the average of recall scores for each class. Area under curve of the receiver operating characteristic (shorthand: ROC-AUC) measures the performance as discrimination threshold of the classifier varies. We set the "multi_class" parameter in the implementation as "ovo", which stands for one-vs-one and calculates the average area under curve (AUC) for all pairs of classes.

Supplementary table 6 shows the results of cross-validation on the terpenes training data. Since LightGBM and random forest performed similarly on producing the best results, we optimized hyperparameters of both methods, employing randomized search, and cross-validated them again on the training data. Carrying out with the optimized hyperparameters (Supplementary table 7), it is shown in Supplementary table 8 that LightGBM performed better, however slightly, while running 4.7 times faster. It reaches to 91% balanced accuracy and 99% ROC-AUC on the validation data. Based on what we described, LightGBM was chosen to classify the terpenes subclasses.

In **Figure 8**, the evaluation results of applying LightGBM on the terpenes data is depicted. **Figure 8**a shows that we approached at least 0.91 (out of 1.00) by all the considered metrics; for instance, balanced accuracy and weighted F1 score on the test data are 0.92 and 0.93, respectively. Also, **Figure 8**b shows the confusion matrix obtained by applying LightGBM on the test data, from which we could calculate the weighted metrics shown in **Figure 8**a, since it denotes true positives, false positives, true negatives and false negatives for each subclass. Given the positive results obtained, we succeeded in applying a classifier capable of classification of terpene molecules on the grounds of their physico-chemical properties.

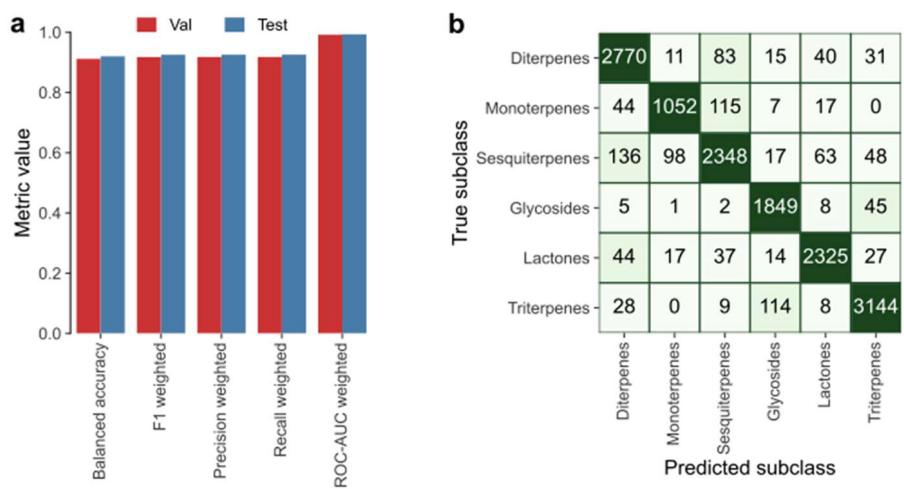

**Figure 8.** LightGBM applied on the terpenes data. (a) Cross-validation and test results and (b) confusion matrix for the test data.

## Conclusion

In this work, we have exploited the significant growth in terms of data accessibility in the field of natural products to better define the chemical space of terpenes. We provide information on a subclass-by-subclass basis of some of the most important parameters for drug development and further contribute to the phytochemical knowledge of this group of natural molecules.

We also tried to apply a number of dimensionality reduction and clustering algorithms to assess the suitability of the available data to this end. The results were not as good as desirable, which calls for increasingly well-annotated data on natural products and new methods.

This work also shows that several algorithms are able to successfully classify terpenes solely using their physico-chemical descriptors and provide information on the performance of said algorithms. The results provided here can be useful for further studies on this class of natural products, including those involving the selection of molecules with favorable characteristics from a pharmaceutical point of view.

# Bibliography

BECHT, E., MCINNES, L., HEALY, J., DUTERTRE, C.-A., KWOK, I. W. H., NG, L. G., GINHOUX, F. & NEWELL, E. W. 2019. Dimensionality reduction for visualizing single-cell data using UMAP. *Nature Biotechnology,* 37**,** 38-44.

BENET, L. Z., HOSEY, C. M., URSU, O. & OPREA, T. I. 2016. BDDCS, the Rule of 5 and drugability. *Advanced Drug Delivery Reviews,* 101**,** 89-98.

BREIMAN, L. 1996. Bagging predictors. *Machine learning,* 24**,** 123-140.

DJOUMBOU FEUNANG, Y., EISNER, R., KNOX, C., CHEPELEV, L., HASTINGS, J., OWEN, G., FAHY, E., STEINBECK, C., SUBRAMANIAN, S., BOLTON, E., GREINER, R. & WISHART, D. S. 2016. ClassyFire: automated chemical classification with a comprehensive, computable taxonomy. *Journal of Cheminformatics,* 8**,** 61.

FIX, E. & HODGES, J. L. 1989. Discriminatory analysis. Nonparametric discrimination: Consistency properties. *International Statistical Review/Revue Internationale de Statistique,* 57**,** 238-247.

GUIMARÃES, A. G., SERAFINI, M. R. & QUINTANS-JÚNIOR, L. J. 2014. Terpenes and derivatives as a new perspective for pain treatment: a patent review. *Expert Opinion on Therapeutic Patents,* 24**,** 243-265.

HO, T. K. Random decision forests. Proceedings of 3rd international conference on document analysis and recognition, 1995. IEEE, 278-282.

HYVÄRINEN, A. & OJA, E. 2000. Independent component analysis: algorithms and applications. *Neural networks,* 13**,** 411-430.

KE, G., MENG, Q., FINLEY, T., WANG, T., CHEN, W., MA, W., YE, Q. & LIU, T.-Y. 2017. Lightgbm: A highly efficient gradient boosting decision tree. *Advances in neural information processing systems,* 30**,** 3146-3154.

LU, J., CHEN, L., YIN, J., HUANG, T., BI, Y., KONG, X., ZHENG, M. & CAI, Y.-D. 2016. Identification of new candidate drugs for lung cancer using chemical–chemical interactions, chemical–protein interactions and a K-means clustering algorithm. *Journal of Biomolecular Structure and Dynamics,* 34**,** 906-917.

MADUGULA, S. S., JOHN, L., NAGAMANI, S., GAUR, A. S., POROIKOV, V. V. & SASTRY, G. N. 2021. Molecular descriptor analysis of approved drugs using unsupervised learning for drug repurposing. *Computers in Biology and Medicine,* 138**,** 104856.

MCINNES, L., HEALY, J. & MELVILLE, J. 2018. Umap: Uniform manifold approximation and projection for dimension reduction. *arXiv preprint arXiv:1802.03426*.

PAWAR, S., LIEW, T. O., STANAM, A. & LAHIRI, C. 2020. Common cancer biomarkers of breast and ovarian types identified through artificial intelligence. 96**,** 995-1004.

RAND, W. M. 1971. Objective criteria for the evaluation of clustering methods. *Journal of the American Statistical association,* 66**,** 846-850.

RINGNÉR, M. 2008. What is principal component analysis? *Nature Biotechnology,* 26**,** 303-304.

ROSENBERG, A. & HIRSCHBERG, J. V-measure: A conditional entropy-based external cluster evaluation measure. Proceedings of the 2007 joint conference on empirical methods in natural language processing and computational natural language learning (EMNLP-CoNLL), 2007. 410-420.

ROUSSEEUW, P. J. 1987. Silhouettes: a graphical aid to the interpretation and validation of cluster analysis. *Journal of computational and applied mathematics,* 20**,** 53-65.

SCHÖLKOPF, B., SMOLA, A. & MÜLLER, K.-R. 1998. Nonlinear component analysis as a kernel eigenvalue problem. *Neural computation,* 10**,** 1299-1319.

SOROKINA, M., MERSEBURGER, P., RAJAN, K., YIRIK, M. A. & STEINBECK, C. 2021. COCONUT online: Collection of Open Natural Products database. *Journal of Cheminformatics,* 13**,** 2.

SOROKINA, M. & STEINBECK, C. 2019. NaPLeS: a natural products likeness scorer—web application and database. *Journal of Cheminformatics,* 11**,** 55.

SUD, P., SU, M. K., GRELLER, H. A., MAJLESI, N. & GUPTA, A. 2013. Case series: inhaled coral vapor--toxicity in a tank. *Journal of medical toxicology : official journal of the American College of Medical Toxicology,* 9**,** 282-286.

VAN DER MAATEN, L. & HINTON, G. 2008. Visualizing data using t-SNE. *Journal of machine learning research,* 9.